\PassOptionsToPackage{table}{xcolor}
\documentclass[runningheads]{llncs}
\usepackage[final,year=2024,ID=201]{eccv}
\usepackage{eccvabbrv}

\usepackage[accsupp]{axessibility}
\usepackage{bbm}
\usepackage{booktabs}
\usepackage{diagbox}
\usepackage{easy-todo}
\usepackage{graphicx}
\usepackage{makecell}
\usepackage{multirow}
\usepackage{xspace}

\usepackage[breaklinks,colorlinks,citecolor=eccvblue]{hyperref}
\usepackage{orcidlink}

\definecolor{indigo}{RGB}{75,0,130}
\definecolor{darkyellow}{RGB}{204, 204, 0}
\definecolor{darkgreen}{RGB}{0, 100, 0}

\newcommand{\method}{DragAPart\xspace}
\newcommand{\dataset}{Drag-a-Move\xspace}

\newcommand{\ipoke}{\textcolor{orange}{iPoke}\xspace}
\newcommand{\dragdiffusion}{\textcolor{darkyellow}{DragDiffusion}\xspace}
\newcommand{\dragondiffusion}{\textcolor{darkgreen}{DragonDiffusion}}\xspace
\newcommand{\dragNUWA}{\textcolor{blue}{DragNUWA}\xspace}
\newcommand{\motionguidance}{\textcolor{indigo}{Motion Guidance}\xspace}
\newcommand{\iptop}{\textcolor{purple}{InstructPix2Pix}\xspace}

\newcommand\rurl[1]{%
\href{https://#1}{\nolinkurl{#1}}%
}

\makeatletter
\renewcommand{\paragraph}{%
  \@startsection{paragraph}{4}%
  {\z@}{-0.5em}{-0.5em}%
  {\normalfont\normalsize\it}%
}
\makeatother

\title{\method: Learning a Part-Level \texorpdfstring{\\}{ } Motion Prior for Articulated Objects}

\author{Ruining Li \and
Chuanxia Zheng \and
Christian Rupprecht \and
Andrea Vedaldi}
\authorrunning{R.~Li et al.}
\institute{Visual Geometry Group, University of Oxford\\
\email{\{ruining, cxzheng, chrisr, vedaldi\}@robots.ox.ac.uk}\\
\rurl{dragapart.github.io}}

\begin{document}
\maketitle

\begin{center}
\includegraphics[trim={0px 0px 0px 0px}, clip, width=\linewidth]{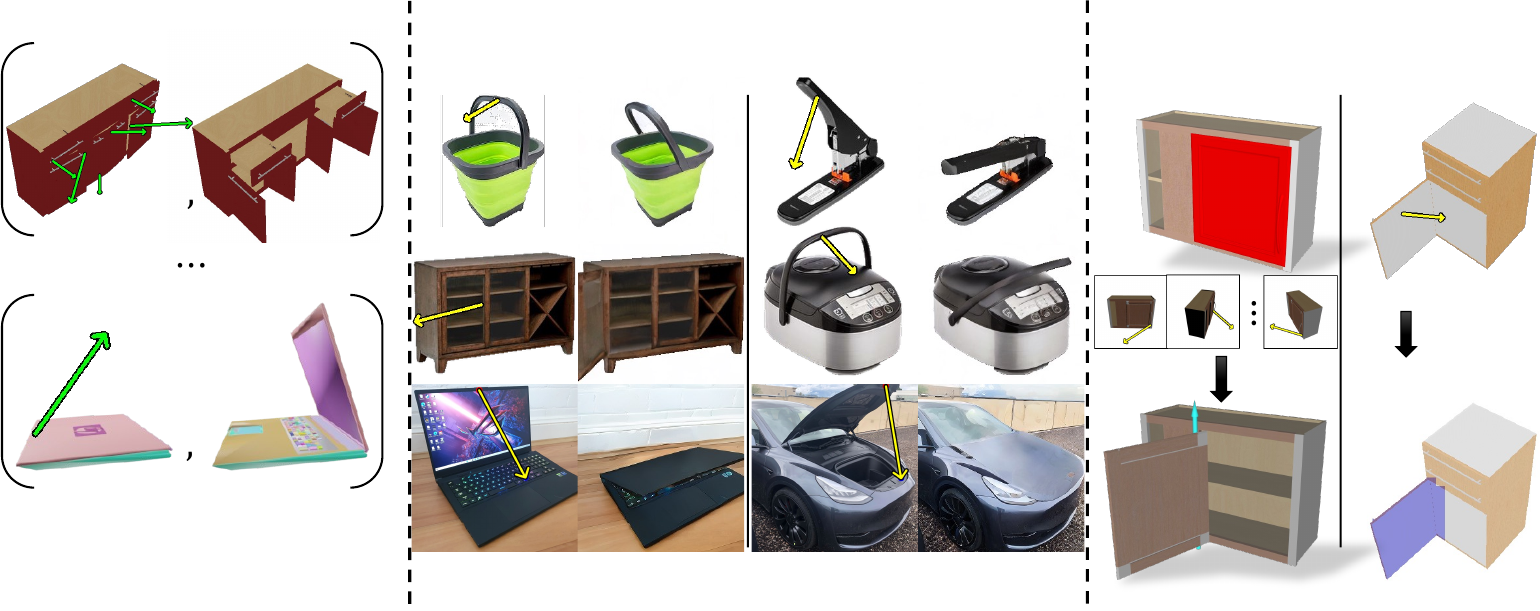}
\begin{picture}(0,0)
    \put(-160,145){\small \textbf{Training Data}}
    \put(-20,145){\small \textbf{Inference}}
    \put(100,145){\small \textbf{Applications}}
    \put(-66,135){\scriptsize	 Seen Categories}
    \put(2,135){\scriptsize \textcolor{red}{Unseen} Categories}
    \put(90,135){\scriptsize Motion}
    \put(88,128){\scriptsize Analysis}
    \put(135,135){\scriptsize Moving Part}
    \put(134,128){\scriptsize Segmentation}
    \put(155,80){\scriptsize Latent}
    \put(153,72){\scriptsize Feature}
    \put(150,64){\scriptsize Clustering}
    \put(-163,25){\scriptsize Synthetic Renderings}
    \put(-171,16){\scriptsize with Sparse Motion Field}
    \put(-72,15){\scriptsize A Motion Prior for Articulated Objects}
\end{picture}
\vspace{-1.7em}
\captionof{figure}{
\textbf{Examples of our \method.}
% Differently from prior works that used drags to move an object,
Each drag in \method represents a \emph{part-level} interaction, resulting in a physically plausible deformation of the object shape.
\method is trained on a new synthetic dataset, \dataset, for this task, and generalizes well to \emph{real} data and even \emph{unseen categories}.
The trained model also can be used for segmenting movable parts and analyzing motion prompted by a drag.
}%
\label{fig:intro}
\end{center}
\begin{abstract}
We introduce \method, a method that, given an image and a set of drags as input, generates a new image of the same object that responds to the action of the drags.
Differently from prior works that focused on repositioning objects, \method predicts part-level interactions, such as opening and closing a drawer.
We study this problem as a proxy for learning a generalist motion model, not restricted to a specific kinematic structure or object category.
We start from a pre-trained image generator and fine-tune it on a new synthetic dataset, \dataset, which we introduce.
Combined with a new encoding for the drags and dataset randomization, the model generalizes well to real images and different categories.
Compared to prior motion-controlled generators, we demonstrate much better part-level motion understanding.
\end{abstract}
\section{Introduction}%
\label{sec:intro}

We consider the problem of learning an \emph{interactive} image generator that allows moving the parts of an object by dragging them~\cite{teng2023drag,pan2023drag} instead of repositioning the object as a whole.
For example, dragging on the door of a cabinet should result in the image of the same cabinet but with the door open (\cref{fig:intro}).

Besides applications to controlled image generation, we explore dragging as a way of learning and probing generalist models of motion.
Modelling deformable objects often uses ad-hoc models specific to each object type.
For example, SMPL~\cite{loper15smpl} only represents humans and SMAL~\cite{zuffi173d-menagerie} only certain mammals.
Some works~\cite{wu2023magicpony,jakab2024farm3d,wu2023dove,li2024learning} have suggested training similar models automatically from raw Internet images but still assuming that objects share the same kinematic structure (a skeleton) which limits generality.
In stark contrast, foundation models like CLIP~\cite{radford2021learning}, GPT-4~\cite{openai23gpt4}, DALL-E~\cite{ramesh21zero-shot} and Stable Diffusion (SD)~\cite{rombach2022stablediffusion} take a generalist approach.
They do not pose any restriction on the type of content in the images or text.
Likewise, a foundation model of motion should be able to understand the motion of any object, whether it is a human, a jellyfish, a towel, or a piece of furniture.
However, there is no universal deformable template that can express the pose and deformation of all objects in a homogeneous manner --- for example, the pose of a human is hardly comparable to that of a cabinet.

We posit that a model of motion does \emph{not} require to refer to a template; it is enough that the model understands the possible physical configurations of an object and their transitions.
Dragging provides a way to probe such a model without using a template.
A drag specifies how a single physical point of the object moves, leaving the model to `fill in' the details by predicting a plausible motion for the object's parts.
For example, dragging a single drawer in a cabinet should not result in all other drawers moving together because their motion is independent, but dragging the handle of a microwave oven's door should open the door itself, as the motion is, in this case, perfectly correlated (\cref{fig:intro}).

The dragging approach also lends itself to large-scale training, potentially from millions of Internet images and videos, by exploiting the strengths of recent generative models~\cite{ramesh21zero-shot,rombach2022stablediffusion,singer22make-a-video, girdhar2023emu}.
In fact, it is natural to ask whether existing off-the-shelf image generators~\cite{ramesh21zero-shot,rombach2022stablediffusion} might already understand dragging and thus motion.
Like others~\cite{shi2024dragdiffusion,mou2023dragondiffusion,teng2023drag,geng2024motion}, we found that these models do respond to dragging, but only in the sense of shifting an entire object, but \emph{fail to capture more nuanced effects}.
Hence, the challenge is to encourage generative models to produce fine-grained and, thus informative dragging outcomes.

Our first contribution is to develop \dataset, a dataset to align image generators with part-level dragging.
Previous works~\cite{shi23mvdream,liu23zero-1-to-3,li2024instant3d,zheng2023free3d} have proposed to use synthetic 3D objects~\cite{deitke22objaverse,deitke23objaverse-xl} to fine-tune pre-trained generative models for new view synthesis (NVS).
Because the base model is trained on billions of images, a relatively small number of such objects, in the order of millions, is sufficient to train generalizable NVS\@.
We thus propose to take a similar approach and contribute data rendered from an existing 3D dataset with rich part-level annotations~\cite{geng2023gapartnet}.
Augmented with drag annotations, this data can be used to fine-tune the generator and learn non-trivial drag responses.
Even so, our synthetic data is far less abundant than that available for NVS, which includes several million objects~\cite{deitke22objaverse,deitke23objaverse-xl}.
Instead, our data is in the hundreds of different synthetic 3D objects due to the cost of annotating parts and kinematics.
While the number of objects is small, we show that \emph{texture randomization} can significantly improve generalization to out-of-domain cases, including real data and unseen categories at test time.

We then use our dataset \dataset to train a new model, \method, for the task of part-oriented drag control.
Key to the design of \method is how the drags are encoded.
Compared to prior works that considered similar tasks, we propose a new way of encoding drags with benefits that are transversal to the underlying model architecture.
We explore different network architectures for this task: DiT~\cite{peebles2023scalable} with a transformer~\cite{vaswani2017attention} backbone trained from scratch and Latent Diffusion architecture~\cite{rombach2022stablediffusion} with a U-Net~\cite{ronneberger15u-net} backbone pre-trained on billions of images, and study their performance on synthetic and real data.

Finally, we also explore some downstream applications of~\method.
First, we show that it can be used to optimize the motion parameters of a given articulated 3D object to predict \emph{how} its movable part is likely to move, subject to a drag.
Second, we show that the model's implicit part-level understanding can be leveraged to segment moving parts in an image prompted by a drag.

\section{Related Work}%
\label{sec:related}

\paragraph{Generative Modeling.}

Generative models, such as diffusion models~\cite{ho2020denoising,song2019generative,song2021scorebased}, can generate high-quality images~\cite{ramesh21zero-shot, rombach2022stablediffusion, saharia2022photorealistic}, videos~\cite{ho2022imagen-video,blattmann2023stable} and 3D assets~\cite{tevet2022human, lei2023nap, liu24cage}.
Conditioning plays a key role in controlling the generated content and its quality.
For instance, DALL-E 3~\cite{betker2023improving} emphasizes the importance of quality textual prompts used for training.
ControlNet~\cite{zhang2023adding} shows how to inject new kinds of controls in a pre-trained diffusion model and is at the basis of a new model zoo~\cite{von-platen-etal-2022-diffusers}, widely used in applications such as text-to-image/video/3D~\cite{ruiz2022dreambooth,guo2023animatediff,poole2022dreamfusion} and image-to-image/video/3D~\cite{hu2023cocktail,zheng2023free3d,watson2023novel,jakab2024farm3d}.

\paragraph{Drag-Conditioned Image \& Video Synthesis.}

Authors have considered controlling image and video generation via dragging.
\emph{Training-free} methods iteratively update the source image to match user-specified drags.
Among those, DragGAN~\cite{pan2023drag} optimizes a latent representation of the image in StyleGAN~\cite{karras2019style} to match the user-specified drags.
DragDiffusion~\cite{shi2024dragdiffusion} ports this idea to SD~\cite{rombach2022stablediffusion}.
DragonDiffusion~\cite{mou2023dragondiffusion} and Motion Guidance~\cite{geng2024motion} combine SD with a guidance term that captures feature correspondences and a flow loss.
\emph{Training-based} methods, on the other hand, learn drag-based control using ad-hoc training data for this task.
For instance, iPoke~\cite{blattmann2021ipoke} trains a variational autoencoder (VAE) to synthesize videos with objects in motion.
MCDiff~\cite{chen2023motion} and YODA~\cite{davtyan2023learn} train a diffusion model using DDPM and flow matching, respectively.
Li \etal~\cite{li2023generative} use a Fourier-based representation of motion suitable for natural, oscillatory dynamics characteristic of objects like trees and candles and generates motion with a diffusion model.
DragNUWA~\cite{yin2023dragnuwa} and MotionCtrl~\cite{wang2023motionctrl} extend text-to-video generators with drag-based control.

Many of these methods show impressive results in terms of repositioning objects in a scene, but \emph{do not} allow to control motion at the level of object parts.
Here, we address the latter task by introducing a new synthetic dataset and a new encoding technique for the drags.

\section{Method}%
\label{sec:method}

We develop \emph{\method}, an \emph{interactive} generative model that, given as input a single object-centric RGB image $y$ and one or more drags $\mathcal{D}$, synthesizes a second image $x \sim \mathbb{P}(x|y, \mathcal{D})$ that reflects the effect of the drags.
We first provide a formal definition of the problem and discuss the challenges of learning such a motion prior for articulated objects in \cref{sec:preliminaries}.
To overcome these challenges, we introduce a novel mechanism to encode the drags $\mathcal{D}$ (\ie, the sparse motion condition), which enables more efficient information propagation in \cref{sec:architecture}.
Since we train our model only on synthetic renderings, in \cref{sec:generalization}, we propose a simple yet effective way to mitigate the sim-to-real gap.
In \cref{sec:applications}, we suggest downstream applications of \method.

\subsection{Preliminaries}%
\label{sec:preliminaries}

We consider images $x,y\in \mathbb{R}^{3 \times \Omega}$ as tensors defined on the spatial grid
$
\Omega = \{1,\dots,H\} \times \{1,\dots,W\}.
$
A \emph{drag} $d$ is a pair $(u,v)\in\Omega\times \mathbb{Z}^2$.
Given the image $y$ and one or more drags
$
\mathcal{D} \subset \Omega \times \mathbb{Z}^2
$,
the goal is to draw samples from the conditional distribution $\mathbb{P}(x|y,\mathcal{D})$.
The origin $u$ of the drag is always contained in the domain $\Omega$ of image $y$, but the termination $v$ may not be since the selected point may move outside the camera's field of view.

The distribution $\mathbb{P}(x|y,\mathcal{D})$ is defined by the expected effect of the drags.
Each image contains a certain object, such as a cabinet or a microwave oven, and the effect of each drag is to interact with the corresponding object part.
Formally, let $M \subset \mathbb{R}^3$ be the (unknown) surface of the 3D object contained in the image and consider a drag $(u,v) \in \Omega\times \mathbb{Z}^2$.
Let $p \in M$ be the physical object point that corresponds to the drag origin $u$.
Then, the effect of the drag $(u,v)$ is to cause the object to deform such that point $p$ moves to a new location $q\in \mathbb{R}^3$ that re-projects as close as possible to the drag termination $v$, while obeying the physical rules that govern the object's deformation.
The effect of multiple drags is to specify multiple such constraints, which must be satisfied simultaneously.
Because there is no explicit model of the 3D shape or kinematic properties of the object in the image, the model must learn these properties implicitly.

Note that not all drags are physically plausible, particularly if the drags are given by a user, so they may not be perfectly physically consistent with respect to the underlying 3D object's geometry.
Hence, the model must simulate a motion of the object that corresponds to the drags as well as possible, but \emph{without breaking physical plausibility}.
On the other hand, drags often under-specify the motion, which justifies using a stochastic rather than deterministic prediction model.
In particular, a single drag can often be satisfied in a physically plausible manner by translating the object as a whole.
This solution is not very informative about the object's dynamics.
Instead, we are interested in \emph{modelling nuanced physical interactions at the level of the individual object parts}.

Such nuanced interactions are typical in natural data ---
for example, dragging on the handle of a drawer normally results in opening the drawer instead of translating the entire cabinet --- so, in principle, should be learned automatically from, \eg, raw Internet data.
Perhaps surprisingly, however, we found that this is \emph{not} necessarily the case.
Following the lead of prior motion-conditioned image generators~\cite{pan2023drag, shi2024dragdiffusion, mou2023dragondiffusion, geng2024motion}, the obvious approach is to start from a pre-trained foundation model like Stable Diffusion~\cite{rombach2022stablediffusion}, and then modify its internal representation of the image $y$ to respond to drags.
The advantage is that foundation models are trained on billions of Internet images and, as such, should have an understanding of almost any object and its configurations.
However, we found that such methods tend to \emph{move the object as a whole} at best, which is uninteresting for us.
We hypothesize that the reason is that the training data of foundation models, while abundant in quantity, inevitably entangles multiple factors such as camera viewpoint and multi-object interaction, making it difficult to understand different \emph{part-level} object configurations.

\subsection{\method: Architecture and Drag Encoding}%
\label{sec:architecture}

Given the challenges identified in \cref{sec:preliminaries}, we now describe \method, which learns a motion prior for articulated objects from a pre-trained image generator.
Key to our model is to fine-tune the motion generator, which models the distribution $\mathbb{P}(x|y, \mathcal{D})$, on a synthetic dataset of triplets $(x,y, \mathcal{D})$.
This dataset will be described in \cref{sec:dataset}.
In addition to this data, which is not very large compared to datasets used to train image generators, we must also address the challenge of generalization.
The rest of this section describes our architectural choices which improve the model's generalizability, both \emph{across domains} to real-world images and \emph{across categories} to those that are not present in the dataset.

\begin{figure}[t]
\includegraphics[trim={0px 0px 0px 0px}, clip, width=\linewidth]{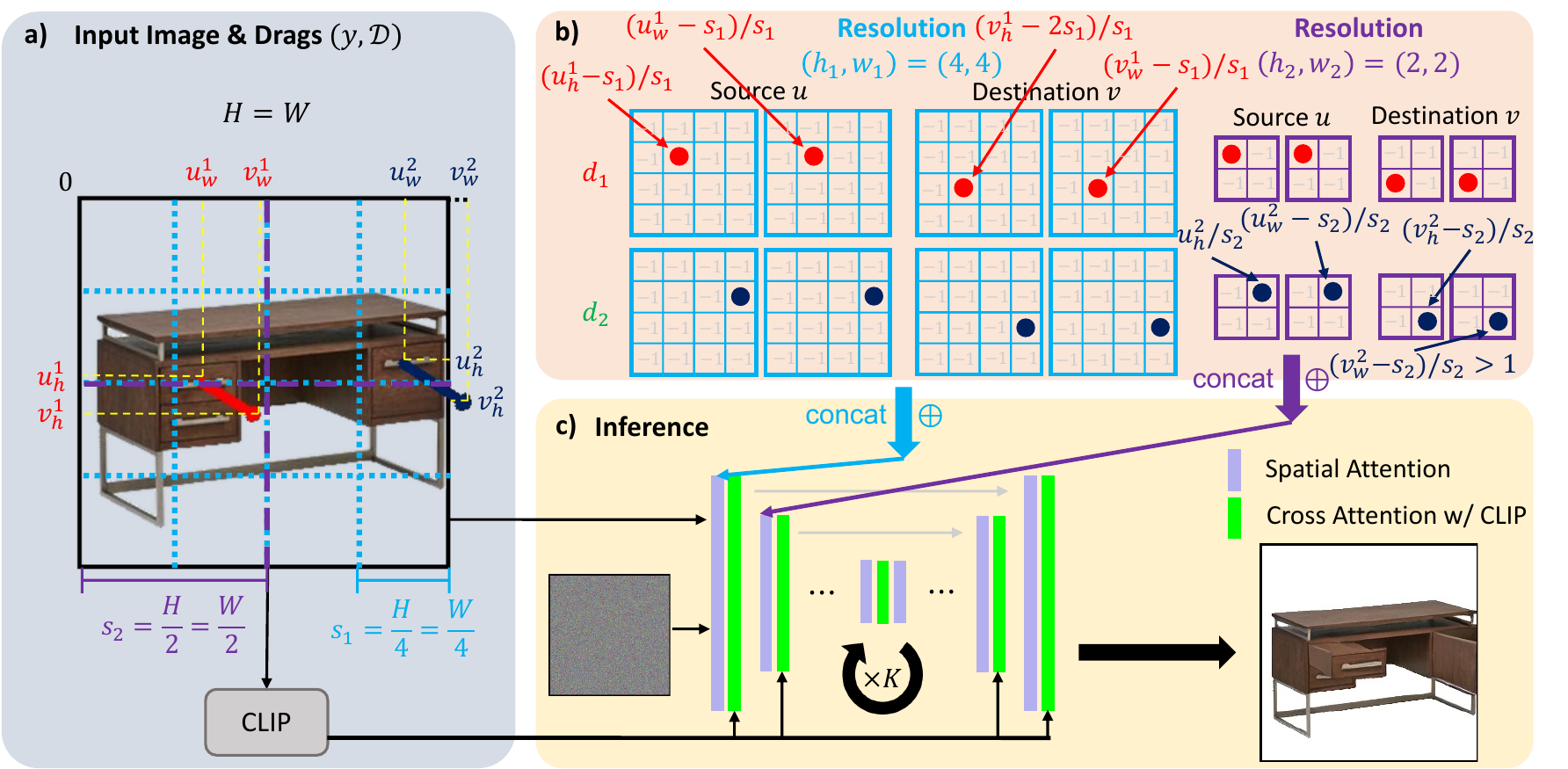}
\vspace{-1.5em}
\captionof{figure}{
\textbf{The Overall Pipeline of \method.}
(a) Our model takes as input a single RGB image $y$ and one or more drags $\mathcal{D}$, and generates a second image $x$ that reflects the effect of the drags (\cref{sec:preliminaries}). (b) We propose a novel flow encoder (\cref{sec:arc_drag_conditions}), which enables us to inject the motion control into the latent diffusion model at different resolutions more efficiently (The resolutions $4$ and $2$ are for illustrative purposes. Our model generates $256\times 256$ images, and the first two latent blocks have resolutions $32$ and $16$.). (c) At inference time, our model generalizes to real data, synthesizing physically-plausible part-level dynamics.
}%
\label{fig:architecture}
\end{figure}

\paragraph{Latent Diffusion Architecture.}

We build our model on large pre-trained diffusion models for image generation, such as Stable Diffusion (SD)~\cite{rombach2022stablediffusion}.
These models are trained on a very large number of natural images (\ie LAION~\cite{schuhmann2022laion}) and are open-ended.
However, they are controlled by textual prompts, whereas our conditioning input consists of a reference image $y$ and a set of drags $\mathcal{D}$.
Hence, our goal is to modify the model to accept the pair $(y, \mathcal{D})$ as conditions.

In more detail, \method uses a Latent Diffusion Model (LDM) architecture and is equipped with an encoder $E$, a decoder $D$ and a denoiser $\Phi$, all implemented as neural networks.
For training, the image $x$ is first mapped to a latent code $z = E(x)$ to which we add Gaussian noise
$
\epsilon \sim \mathcal{N}(0, \mathbf{I})
$
to obtain
$
 z_t = \sqrt{1-\sigma_t^2} z + \sigma_t \epsilon
$,
where $\sigma_0,\dots,\sigma_T$ is a noise schedule that increases monotonically from 0 to 1.
For efficiency, the code $z \in \mathbb{R}^{4\times h \times w}$ is designed to have reduced resolution $(h,w)$ compared to the image resolution $(H,W)$.
The denoising model $\Phi$ is trained by optimizing  the objective function
\begin{equation}\label{eq:objective}
\min_\Phi
\mathbb{E}_{(x,y,\mathcal{D}), t, \epsilon\sim\mathcal{N}(0,1)}
\left[
\lVert
\epsilon - \Phi(z_t, t, y,\mathcal{D})
\rVert^2_2
\right].
\end{equation}

For the network $\Phi$, we experiment with two diffusion architectures, SD and DiT.
The SD model is a UNet with several transformer blocks $l=1,2,\dots,L$ operating at different spatial resolutions $(h_l,w_l)$.
It is conditioned on a textual prompt encoded using CLIP~\cite{radford2021learning}.
At each UNet block $l$, a self-attention layer generates spatial queries, keys and values $Q_l,K_l,V_l \in \mathbb{R}^{C_l\times h_l\times w_l}$ at resolution $(h_l,w_l)$.
In order to inject the source image $y$ as a condition, we follow~\cite{cao_2023_masactrl, weng2023consistent123} and pass $y$ (without adding noise to it) through the same network $\Phi$.
Then, the keys and values $K_l^y,V_l^y$ from image $y$ are used to replace the keys and values $K_l,V_l$ for the generated image $x$, thus turning self-attention into cross attention between the noised image and the clean reference image $y$.
In order to encode the drags $\mathcal{D}$, we consider a drag encoder $F_l(\mathcal{D}) \in \mathcal{R}^{C_l\times h_l\times w_l}$ outputting an encoding of the drags at the resolution compatible with transformer block $l$.
These features are concatenated along the channel dimension to the input of each transformer block.
The transformer layers are modified to handle the additional input channels, zero-initializing the corresponding model weights.

The DiT model has several attention blocks at the same latent resolution.
For each block, we follow~\cite{peebles2023scalable} and inject the (tokenized) drag encoding via a token-wise adaLN-zero block~\cite{perez2018FiLM}.
We defer the details to the sup.~mat.

\paragraph{Drag Encoding.}%
\label{sec:arc_drag_conditions}

We now define the encoding $F_l(\mathcal{D})$ of the drags.
Prior works~\cite{yin2023dragnuwa, wang2023motionctrl} represent $\mathcal{D}$ as a sparse optical flow image
$
F(\mathcal{D}) \in \mathbb{R}^{2\times \Omega}
$
which is all zero except that, for each drag $(u,v)\in\mathcal{D}$, location $u$ contains the displacement $v - u$.
This mechanism has two shortcomings.
First, the transformer can sense well the origin $u$ of the drag, which is encoded spatially, but cannot do so for the termination $v$.
Second, because all drags are encoded by a single optical flow image and the latter must be reduced to resolution $(h_l,w_l)$ before it can be injected into the LDM transformer block $l$, distinct drags may overlap.
For instance,~\cite{yin2023dragnuwa, wang2023motionctrl} first blur the sparse flow image $F$ using a Gaussian filter and then apply further convolutional layers to the result before injecting the information into the LDM\@.
As demonstrated in \cref{sec:experiments}, this design can confuse different drags and diffuse a drag so that it affects more than the intended part (\cref{fig:ablations} left).

In order to mitigate these downsides, we introduce a new encoding $F_l$ for the drags, illustrated in \cref{fig:architecture}.
First, we assume a maximum number $|\mathcal{D}| \leq N$ of drags and assign different channels to different ones, so that \emph{they do not interfere with each other in the encoding}.
Second, we use \emph{separate channels} to encode the drag source $u$ and the drag termination $v$.
Specifically, given a single drag $d=(u,v)$ where  $u = (u_h, u_w)$, let $F_l(u,v) \in \mathbb{R}^{2\times h_l \times w_l}$ be the tensor of all negative ones\footnote{So we distinguish between no drag and drags precisely between two latent pixels.} except for
$
F_l\left(
    \left\lfloor\frac{u_h\cdot h_l}{H}\right\rfloor,
    \left\lfloor\frac{u_w\cdot w_l}{W}\right\rfloor
\right)
=
\left(
    \frac{u_h\cdot h_l}{H} - \left\lfloor\frac{u_h\cdot h_l}{H}\right\rfloor,
    \frac{u_w\cdot w_l}{W} - \left\lfloor\frac{u_w\cdot w_l}{W}\right\rfloor
\right)
\in [0,1]^2
$.
We then define
$
F_l(\mathcal{D})
=
\bigoplus_{(u,v)\in\mathcal{D}}
F_l(u,v) \oplus F_l(v,u)
\in \mathcal{R}^{4N\times h\times w}
$
to be the concatenation of the encoding of all drags in $\mathcal{D}$ along the channel dimension.
Note that we encode each drag twice, swapping source and termination.\footnote{There is a caveat when the drag termination $v$ is outside the image, in which case
$
\left(
    \left\lfloor\frac{u_h\cdot h}{H}\right\rfloor,
    \left\lfloor\frac{u_w\cdot w}{W}\right\rfloor
\right)
$
is outside resolution $(h, w)$, so we define the termination $v$ at the nearest spatial location \emph{inside} $(h, w)$.
The corresponding values will also be relative to this new coordinate, so they might not lie in $\left[0, 1\right]$ (see \cref{fig:architecture} for an example).
}
If there are less than $N$ drags, we simply pad the tensor with $-1$.
We show in \cref{sec:experiments} that our proposed \emph{multi-resolution drag encoding} technique boosts the generation quality in both U-Net and attention-based LDMs.

\subsection{Generalization to Real Data via Domain Randomization}%
\label{sec:generalization}

Most objects in GAPartNet~\cite{geng2023gapartnet}, which we use to construct the dataset used to train our model (see \cref{sec:dataset}), have synthetic textures not representative of real-world objects.
Since these textures usually lack variability in color, models trained on these renderings can potentially ``cheat'' by generating the same pixel values everywhere without an implicit understanding of where the pixel comes from in the original image $y$.
Empirically, we find training the model jointly on a version of the renderings in which each object part is assigned a random color as shown in \cref{fig:intro} and \cref{fig:dataset} (right), sometimes called \emph{domain randomization}~\cite{tobin17domain}, significantly improves the model's performance on real-world images.
The effectiveness of this training strategy is validated in \cref{sec:experiments}.

\subsection{Downstream Applications}%
\label{sec:applications}

While our primary goal is to demonstrate that an image generator can learn a motion prior for articulated objects at \emph{part-level}, we also explore the potential of our \method in several downstream applications.
By operating at the level of object parts, our approach is complementary to prior works that focus on the task of repositioning~\cite{yin2023dragnuwa, wang2023motionctrl} and can address different applications than them.

\paragraph{Segmenting Moving Parts.}

First, \method can be used to segment the moving parts of an object.
Several prior works~\cite{baranchuk2021label, xu2023open, tian2023diffuse, karazija2023diffusion} have used pre-trained generators for 2D segmentation, but focus on part semantic rather than part mobility.
We suggest that \method, as a motion prior, can be used to segment the part that would move in response to a drag (useful, \eg, for predicting affordances).

In order to segment mobile parts, we follow~\cite{baranchuk2021label} and extract internal features from the forward pass of the denoiser $\Phi$.
However,~\cite{baranchuk2021label} looks at semantics and only extracts a single set of features from the input image.
Instead, because our goal is to obtain the segmentation of the part that moves prompted by a drag, we run the forward pass twice, once with the drag conditioning and once without, and compute the difference between the resulting features.
This difference represents the effect of the drag in the diffusion model's internal representation.

\paragraph{Motion Analysis for Articulated Objects.}

Next, we consider motion analysis, \ie, the problem of understanding how the parts of an object move~\cite{Abbatematteo2019LearningTG, Mo2021Where2ActFP, gadre2021act, jiang2022ditto, geng2023gapartnet, liu2023paris}.
Here, we suggest a method to use \method for motion analysis.
Given a 3D mesh
$
\mathcal{M} = \mathcal{M}_{\text{static}} \cup \mathcal{M}_{\text{moving}}
$
with moving parts pre-segmented and a 3D drag on $\mathcal{M}$, the goal is to output the type of motion $t$ and the  parameters $p_\text{motion}$%
\footnote{%
Following~\cite{liu2023paris} for each part.
The motion type is one of $\left \{ \text{revolute, prismatic}\right \}$.
For a revolute joint, the motion parameter $p_\text{motion}$ is represented by a pivot point $\Vec{p}\in \mathbb{R}^3$ and a rotation in the form of a quaternion $\Vec{q}\in \mathbb{R}^4$; for a prismatic joint, the parameter specifies the joint axis $\Vec{a}\in \mathbb{R}^3$ and translation distance $s \in \mathbb{R}$ along this axis.
}.

In order to estimate the parts' motion types and parameters using \method, we start by rendering $K$ images $y_{k}$ of the object $\mathcal{M}$ from random cameras $C_{k}$. We also project the 3D drag to pixel space $d_{\text{pixel}}^{k} \in \Omega \times \mathbb{Z}^2$.
Then, we use \method to generate images
$
x_{k} \sim \mathbb{P}(x_{k} | y_{k}, \left \{ d_{\text{pixel}}^k \right \})
$
of the dragged object under $C_k$.
Following~\cite{liu2023paris}, using the generated images, we estimate the motion parameters
$
(t, p_\text{motion})
$
by minimizing the objective function
\begin{equation}
\label{e:motion-analysis-objective}
 \operatornamewithlimits{argmin}_{t,\;p_\text{motion}}
    \frac{1}{K}\sum_{k=1}^{K} \lVert
        R\left(\mathcal{M}_\text{static}\cup T\left(\mathcal{M}_\text{moving}; t, p_\text{motion}\right), C_{k}\right) - x_k
    \rVert_2^2
\end{equation}
where $R$ is the rendering function and $T$ transforms the moving parts based on the motion estimation.
While~\cite{liu2023paris} solves~\cref{e:motion-analysis-objective} using back-propagation, we find gradient-based optimization prone to local optima, leading to sub-optimal estimation.
Since the search space has relatively low dimensionality\footnote{$7$ for revolute type and $4$ for prismatic type.},
we optimize~\cref{e:motion-analysis-objective} directly using grid search.

\section{The \dataset~Dataset}%
\label{sec:dataset}

Key to training our model is a suitable dataset with drag annotations.
It must thus contain a collection of triplets $(x,y,\mathcal{D})$,
where image $y$ represents the initial state of the object, $\mathcal{D}$ is the collection of drags applied to $y$
and image $x$ denotes the same object in a new state compatible with the action of the drags.
Moreover, we wish the state change to be minimal, in the sense that each drag only affects the smallest subset of object parts that still result in a physically plausible state.

Creating such a dataset using real-world images would require carefully capturing photos of an articulated object in two articulation states while ensuring consistency in camera position, lighting condition and background across captures, and then manually annotating or reconstructing the drags.
Fortunately, recent works on novel-view synthesis have indicated that generative models can be effectively fine-tuned using synthetic data only.
We thus choose to create a synthetic dataset, which allows us to control the articulation and lighting conditions, and to obtain the ground-truth drags directly from the 3D model.

We build our dataset on an existing 3D synthetic dataset, GAPartNet~\cite{geng2023gapartnet}, which adds rich, part-level annotations to the objects of~\cite{xiang2020sapien,liu22akb-48}.
By rendering the assets under different articulations, we can obtain synthetic data with desired properties, as described next.
\begin{figure}[tb!]
\includegraphics[trim={0px 0px 0px 0px}, clip, width=\linewidth]{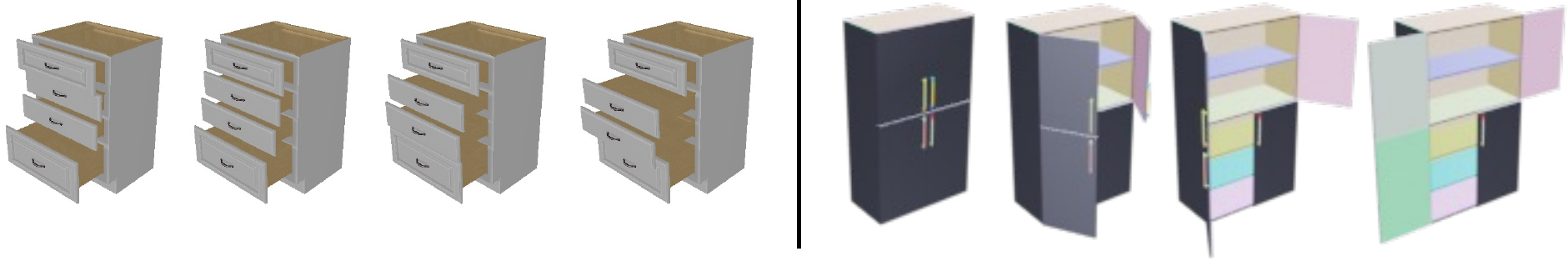}
\vspace{-2.0em}
\caption{
\textbf{Animations from the \dataset dataset.}
We visualize two objects with diverse articulation states: the left is rendered with the original texture and the right with each part in a single random color.
}%
\label{fig:dataset}
\end{figure}
Among all the objects in GAPartNet, we select those that have at least one part which is annotated with one of the labels ``Hinge Handle'', ``Slider Drawer'', ``Hinge Lid'' and ``Hinge Door'', which results in $763$ candidate 3D models spanning $16$ categories.
See the sup.~mat.~for details.

\paragraph{Animating the Assets.}

The annotation of each object contains its kinematic tree structure, where nodes are rigid parts and edges are joints (fixed or not).
Our goal is to sample a diversity of starting and ending states (before and after drag) of the different parts, including cases in which only a subset of parts move.
We generate with equal probability two kinds of animations.
In the first, some joints are fixed and locked to their ``closed'' or ``rest'' state, whereas the other joints transition continuously between two extreme states (\eg, a drawer fully closed or fully open, \cref{fig:dataset} right).
The second kind is similar, but the stationary joints are set to a random state, and the moving joints transition between two randomly selected states (\cref{fig:dataset} left).
This is further detailed in the sup.~mat.

\paragraph{Generating the Drags.}

Since we have ground-truth 3D data with articulation annotation, we can project the motion
$
(p,q) \in \mathbb{R}^3 \times \mathbb{R}^3
$
of any visible 3D point on the moving part back to pixel space to obtain a drag $(u,v)$.
However, this would result in a dense set of drags, akin to optical flow, whereas we are interested in only selecting one drag per part.
In order to generate a sparse subset of drags, for each moving joint we first sample a part in the subtree (of the entire kinematic structure) rooted at that joint uniformly at random.
Then, we sample a point on the visible surface of that part with probabilities proportional to its displacement due to the parent joint's motion.
We use the trajectory of that point as the drag for the moving parts.
This process is fast, enabling us to generate different drag combinations \emph{on the fly} during training to augment the dataset further.

\paragraph{Rendering.}

For each 3D asset, we render $48$ articulation animations, with 36 frames each, under random lighting conditions and camera viewpoints by means of the SAPIEN~\cite{xiang2020sapien} renderer.
This amounts to $40$ million unique image pairs $(x, y)$ for training purposes.
% The images and drags are constructed at resolution $512\times 512$.
We also render a version of the dataset where each part of the 3D model is given a random, monochromatic color texture (\cref{fig:dataset} right), which we find useful for the model's generalizability (\cref{sec:generalization}).
\section{Experiments}%
\label{sec:experiments}

We show that our method outperforms prior works both quantitatively and qualitatively, and conduct ablation studies to validate our design choices.
% Please refer to the sup.~mat.~for additional results and details.

\subsection{Datasets and Implementation Details}

\paragraph{Articulated Objects.}

We assess \method on the test split of our \dataset dataset (\cref{sec:dataset}),
which contains $150$ image pairs from $23$ objects not seen during training spanning $10$ categories.
Furthermore, we also conduct evaluations on real images sourced from the Common Object in 3D (CO3D) dataset~\cite{reizenstein21co3d} and the Amazon Berkeley Objects (ABO) dataset~\cite{collins2022abo}.
For these, we use SAM~\cite{Kirillov_2023_ICCV} for object segmentation and manually define plausible drags.

As there is no dataset of real images for this task to test generalizability, we generate a second version of the test split using Blender and extract the drags using an off-the-shelf point tracker~\cite{karaev2023cotracker}.
This is closer to real-world data because Blender generates more realistic images and the tracker's inherent inaccuracies produce trajectories more representative of the user-defined drags compared to using ground-truth 3D information.
We refer to this version of the test set as Out-of-Distribution (O.O.D.) and the original version as In-Distribution (I.D.).

\paragraph{Humans.}

We train a separate model for humans on the Human3.6M dataset~\cite{h36m_pami}.
Following previous works~\cite{blattmann2021ipoke, chen2023motion}, we use actors S1, S5, S6, S7 and S8 for training and S9 and S11 for testing.
We randomly sample two frames, as the conditioning image and target, within a video sequence.
The corresponding drags are derived using the annotated paths of keypoints that show sufficient displacement.

\paragraph{Implementation Details.}

We evaluate SD~\cite{rombach2022stablediffusion} and DiT~\cite{peebles2023scalable} as LDM architectures.
In our implementation, we minimally alter the original architecture to enable a more accurate comparison of the capacities of both architectures.
Detailed modifications and our training schedule are provided in the sup.~mat.

A notable difference in training between the two architectures is that we fine-tune the U-Net LDM from a pre-trained checkpoint SD v1.5~\cite{rombach2022stablediffusion}, while we train from scratch for DiT~\cite{peebles2023scalable}, due to the lack of publicly available DiT checkpoints trained on Internet-scale data.
We compare the performance of the two architectures on the I.D.~\dataset in \cref{tab:compare-architecture}.
It should be noted that the conditional DiT lacks pre-training, thereby limiting its generalizability to real-world scenarios.
In the rest of this section, unless stated otherwise, our results are obtained with the fine-tuned U-Net LDM checkpoints.

\begin{table}[t]
\small
\centering
\begin{tabular}{@{}lccc c ccc@{}}
\toprule
\multicolumn{1}{c}{\multirow{2}{*}[0ex]{\textbf{Method}}} &
\multicolumn{3}{c}{\textbf{I.D. \dataset}} & &
\multicolumn{3}{c}{\textbf{Human3.6M}~\cite{h36m_pami}} \\
\cline{2-4}
\cline{6-8}
& PSNR$\uparrow$ & SSIM$\uparrow$ & LPIPS$\downarrow$ & & PSNR$\uparrow$ & SSIM$\uparrow$ & LPIPS$\downarrow$ \\
\midrule
\ipoke~\cite{blattmann2021ipoke} & 16.79 & 0.883 & 0.150 && 21.43 & 0.856 & 0.258 \\
\dragdiffusion~\cite{shi2024dragdiffusion} & 15.30 & 0.773 & 0.226 && 18.06 & 0.707 & 0.280 \\
\dragondiffusion~\cite{mou2023dragondiffusion} & 17.63 & 0.852 & 0.162 && 19.45 & 0.765 & 0.258 \\
\dragNUWA~\cite{yin2023dragnuwa} & 13.58 & 0.765 & 0.277 && 15.16 & 0.668 & 0.292 \\
\method (Ours) & \textbf{21.38} & \textbf{0.925} & \textbf{0.066} && \textbf{23.82} & \textbf{0.870} & \textbf{0.091} \\
\bottomrule
\end{tabular}
\caption{
\textbf{Quantitative Comparisons} on our \dataset and Human3.6M~\cite{h36m_pami}.
\vspace{-1em}
}%
\label{tab:compare-all}
\end{table}

\subsection{Comparisons with Prior Work}

\paragraph{Baselines.}

We perform a thorough comparison of \method to the state-of-the-art generative models which are applicable to this task:
\ipoke~\cite{blattmann2021ipoke}, \dragdiffusion~\cite{shi2024dragdiffusion}, \dragondiffusion~\cite{mou2023dragondiffusion}, \dragNUWA~\cite{yin2023dragnuwa}, and \motionguidance~\cite{geng2024motion}.
For the training-free methods (\dragdiffusion, \dragondiffusion~and \motionguidance), we use their official implementations.
Since \motionguidance is designed for dense motion manipulation, we include two versions in our comparison, one given a dense motion field as input and the other adapted to take only sparse drags.
We train \ipoke using the publicly available code on our \dataset dataset and use its released checkpoint trained on Human3.6M~\cite{h36m_pami}.
For \dragNUWA, we use the released checkpoint.
As it is a large-scale open-domain video generator, we do not fine-tune it with our data.
\ipoke and \dragNUWA generate videos, from which we take the last frame for comparison.
Additionally, we compare with another naive baseline, \iptop~\cite{brooks2022instructpix2pix}, which fine-tunes SD~\cite{rombach2022stablediffusion} on a dataset of paired images and corresponding text edits generated with a large language model~\cite{brown2020language}.
As it is conditioned on textual instructions, we manually craft prompts to convey the manipulation of the drags.

% \begin{figure}[tb!]
% \centering
% \includegraphics[width=\linewidth]{figures/qualitative-comp.pdf}
% \caption{\textbf{Qualitative Comparisons}
% on real images from the ABO~\cite{collins2022abo} dataset with manually defined drags (a-d) and the Human3.6M~\cite{h36m_pami} dataset (e) and a rendered image from our \dataset dataset (f). The images generated by our model are more realistic and capture nuanced part-level dynamics.}%
% \label{fig:qualitative-comp}
% \end{figure}

% \begin{center}
% \includegraphics[trim={0px 0px 0px 0px}, clip, width=\linewidth]{figures/qualitative-comp.pdf}
% \vspace{-1.5em}
% \captionof{figure}{
% \textbf{Qualitative Comparison with Prior Work.}
% PLACEHOLDER CAPTION.
% }%
% \label{fig:qualitative-comp}
% \end{center}

\begin{figure}[tb!]
\centering
\includegraphics[width=\linewidth]{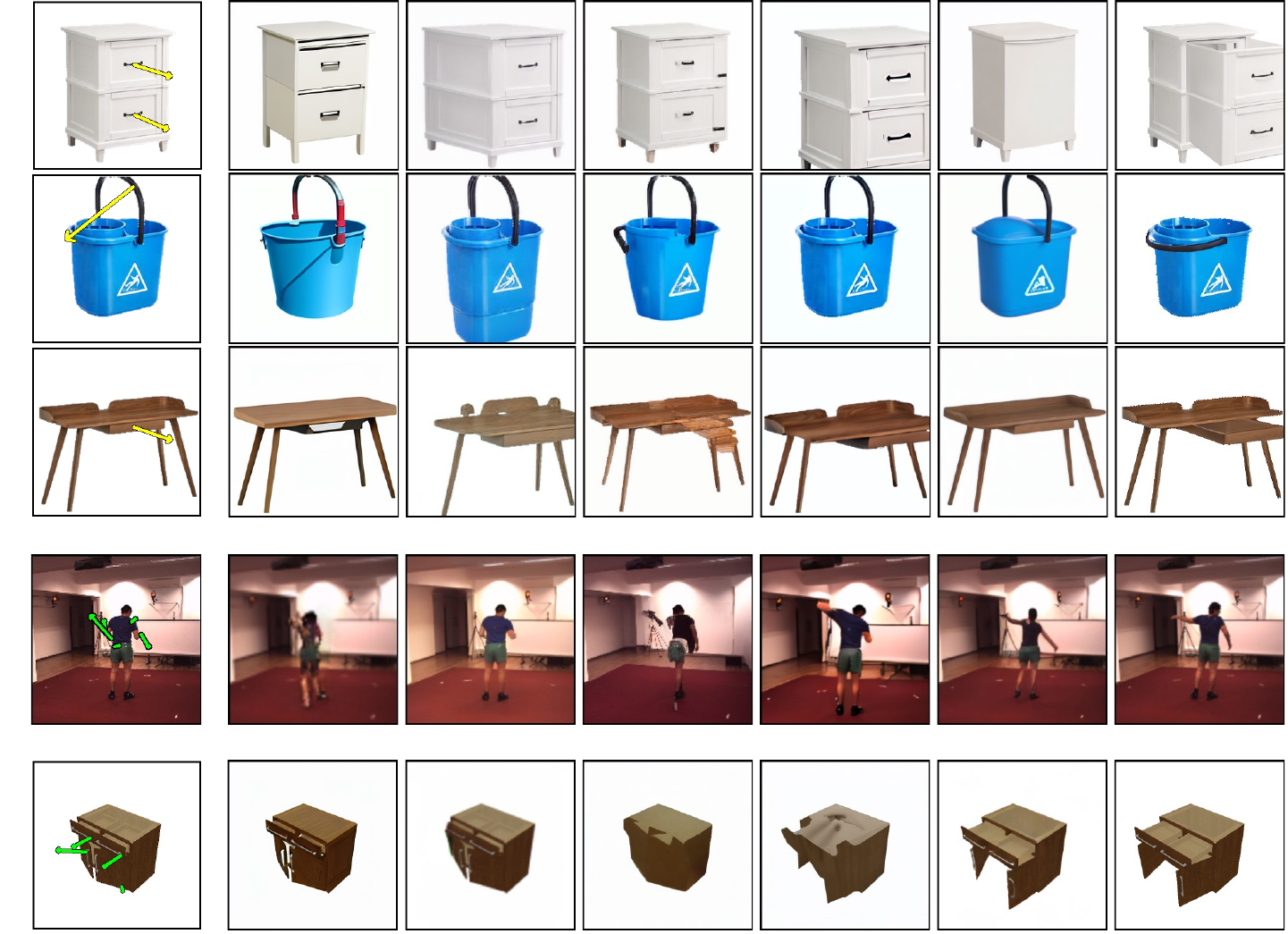}
\begin{picture}(0,0)
    \put(-175,240){\footnotesize a)}
    \put(-175,190){\footnotesize b)}
    \put(-175,145){\footnotesize c)}
    \put(-175,85){\footnotesize d)}
    \put(-175,30){\footnotesize e)}

    \put(-145,263){\footnotesize 0}
    \put(-90,263){\footnotesize 1}
    \put(-44,263){\footnotesize 2}
    \put(5,263){\footnotesize 3}
    \put(52,263){\footnotesize 4}
    \put(100,263){\footnotesize 5}
    \put(148,263){\footnotesize 6}

    \put(-149,118){\tiny \textbf{Input}}
    \put(-118,118){\tiny \textbf{\iptop}}
    \put(-64,118){\tiny \textbf{\dragdiffusion}}
    \put(-18,118){\tiny \textbf{\dragondiffusion}}
    \put(37,118){\tiny \textbf{\dragNUWA}}
    \put(78,118){\tiny \textbf{\motionguidance}}
    \put(145,118){\tiny \textbf{Ours}}

    \put(-149,62){\tiny \textbf{Input}}
    \put(-96,62){\tiny \textbf{\ipoke}}
    \put(-65,62){\tiny \textbf{\dragdiffusion}}
    \put(-19,62){\tiny \textbf{\dragondiffusion}}
    \put(36,62){\tiny \textbf{\dragNUWA}}
    \put(95,62){\tiny \textbf{Ours}}
    \put(128,62){\tiny \textbf{Ground Truth}}

    \put(-149,5){\tiny \textbf{Input}}
    \put(-118,5){\tiny \textbf{\iptop}}
    \put(-50,5){\tiny \textbf{\ipoke}}
    \put(-3,5){\tiny \textbf{\textcolor{indigo}{MG(S)}}}
    \put(45,5){\tiny \textbf{\textcolor{indigo}{MG(D)}}}
    \put(95,5){\tiny \textbf{Ours}}
    \put(128,5){\tiny \textbf{Ground Truth}}
\end{picture}
\vspace{-1em}
\caption{\textbf{Qualitative Comparisons}
on real images from the ABO~\cite{collins2022abo} dataset with manually defined drags (a-c) and the Human3.6M~\cite{h36m_pami} dataset (d) and a rendered image from our \dataset test split (e). The images generated by our model are more realistic and capture nuanced part-level dynamics.
\vspace{-1em}
}%
\label{fig:qualitative-comp}
\end{figure}
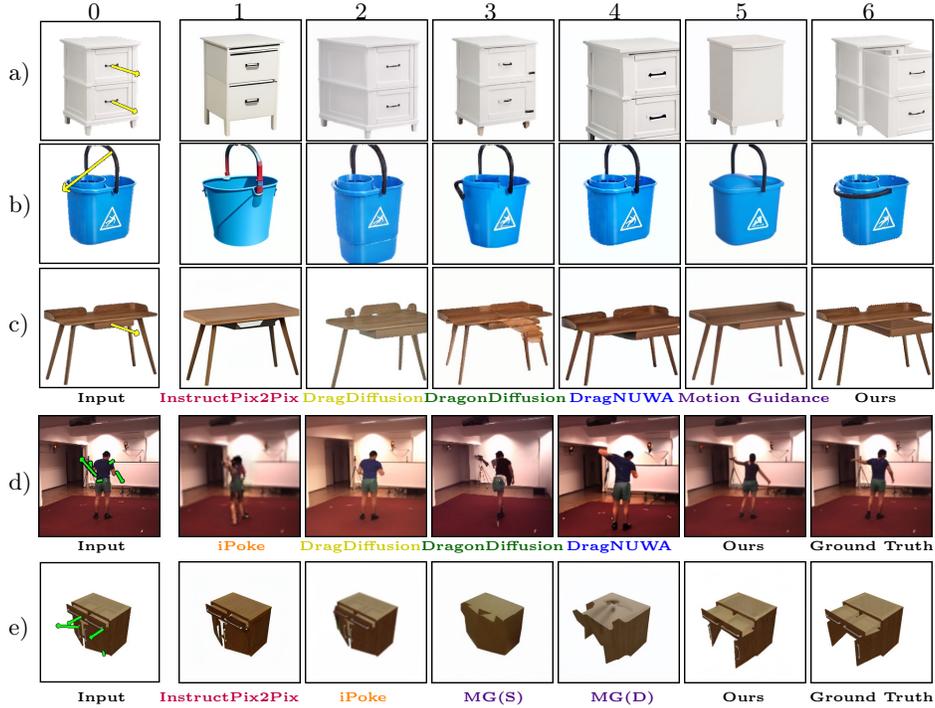

\paragraph{Quantitative Comparisons}

are conducted on the test split of I.D.~\dataset and Human3.6M in \cref{tab:compare-all}.
\method significantly outperforms the state-of-the-art methods on all standard metrics, including pixel-level PSNR, patch-level SSIM, and feature-level LPIPS.
This includes \ipoke~\cite{blattmann2021ipoke}, which has been trained with the identical experimental setting as \method, and \dragNUWA, which is a large pre-trained generative model with motion control, but mainly focuses on moving objects as a whole and does not yield motion resulting from part-level interactions.
\iptop and \motionguidance are excluded in quantitative comparisons because curating a textual instruction for every image pair in the test set is impractical, and we could not run \motionguidance on the entire test set as it requires one hour per instance.
% takes an hour to generate a single image, making it infeasible to run it for every test example in our dataset.

\paragraph{Qualitative Comparisons}

are shown in \cref{fig:qualitative-comp}.
Despite our best efforts, \iptop is never able to deform the object significantly, which motivates using dragging for conditioning instead of language.
Methods that build on Stable Diffusion~\cite{rombach2022stablediffusion} latent features (\dragdiffusion, \dragondiffusion, \motionguidance) tend to copy the appearance of the handle points to the target points without moving the entire part (a-d3, e4).
For \dragdiffusion and \dragNUWA, the object is re-scaled (a2, b2, c4) or repositioned (a4, c2),  rather than performing a \emph{part-level} motion.
For \motionguidance, even when the dense motion field is given (e4), it fails to preserve the object identity very well (a-b5, e4).
By contrast, by using a data-driven approach, our model learns to generate images that preserve faithfully the identity and that exhibit physically plausible motion.
Applied to humans (d), our method produces a result that is closest to the ground truth, without changing the identity of the actor, as opposed to \ipoke (d1).

\paragraph{Qualitative Results on Real Data.}

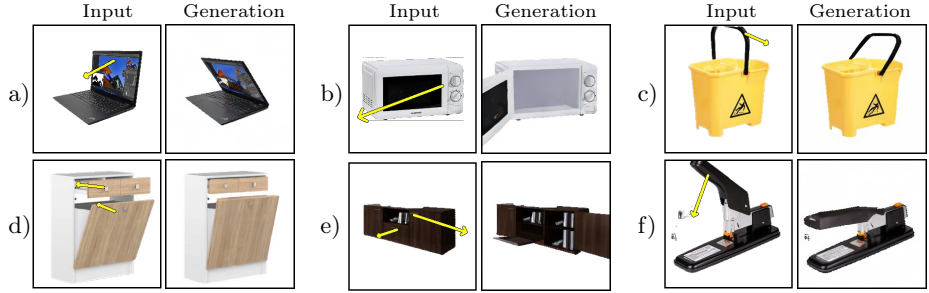
\begin{figure}[tb!]
\includegraphics[trim={0px 0px 0px 0px}, clip, width=\linewidth]{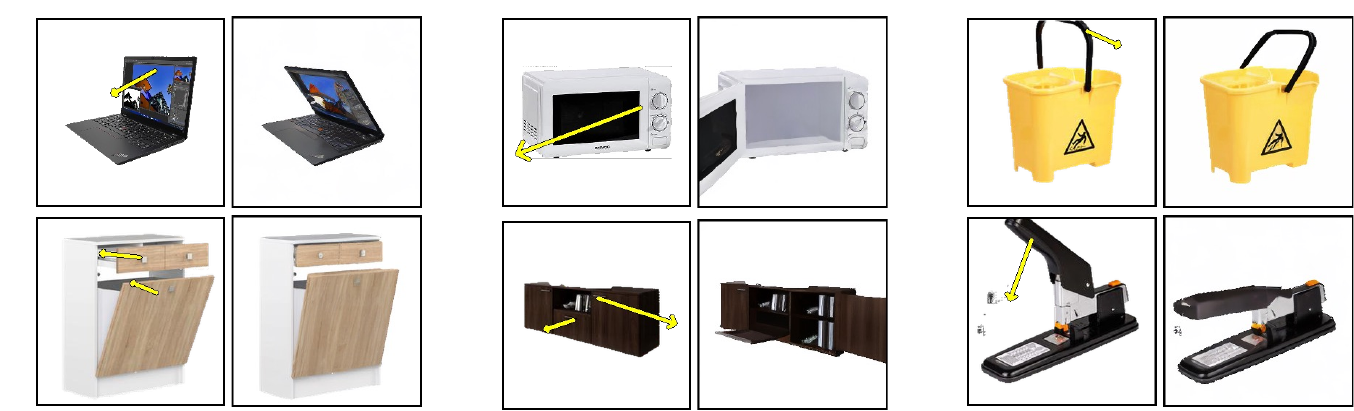}
\begin{picture}(0,0)
    \put(28,115){\scriptsize Input}
    \put(66,115){\scriptsize Generation}
    \put(144,115){\scriptsize Input}
    \put(184,115){\scriptsize Generation}
    \put(264,115){\scriptsize Input}
    \put(303,115){\scriptsize Generation}
    \put(0,83){\footnotesize a)}
    \put(118,83){\footnotesize b)}
    \put(238,83){\footnotesize c)}
    \put(0,33){\footnotesize d)}
    \put(118,33){\footnotesize e)}
    \put(238,33){\footnotesize f)}
\end{picture}
\vspace{-1.5em}
\caption{
\textbf{Qualitative Results on Real Data.} Trained only on synthetic renderings, \method shows excellent generalizability to both \emph{real images} and \emph{novel categories}.
}%
\label{fig:qualitative-results}
\end{figure}

\cref{fig:qualitative-results} shows real images generated by our model, which can
(a) preserve fine-grained texture details,
(b) generate reasonable shades,
(c) handle thin structures,
(d) compose multi-part motion,
(e) ``dream up'' the internal structures of the object,
and
(f) generalize to categories not seen during training.
More results are presented in the sup.~mat.

\begin{table}[t]
\small
\setlength\tabcolsep{4pt}
\renewcommand{\arraystretch}{1.0}
\centering
\begin{tabular}{@{}l ccc c ccc@{}}
\toprule
\multicolumn{1}{c}{\multirow{2}{*}[0ex]{\textbf{Flow Encoding}}} & \multicolumn{3}{c}{\textbf{DiT}~\cite{peebles2023scalable}} && \multicolumn{3}{c}{\textbf{SD}~\cite{ronneberger15u-net, rombach2022stablediffusion}} \\
\cline{2-4}
\cline{6-8}
\multicolumn{1}{c}{} & PSNR$\uparrow$ & SSIM$\uparrow$ & LPIPS$\downarrow$ && PSNR$\uparrow$ & SSIM$\uparrow$ & LPIPS$\downarrow$ \\
\midrule
Conv. Input Only & 19.56 & 0.910 & 0.095 && 19.97 & 0.914 & 0.077 \\
% \midrule
Conv. Every Block & 20.58 & 0.922 & 0.078 && 21.10 & \textbf{0.925} & 0.067 \\
Multi-Res. Enc. Every Block & \textbf{21.11} & \textbf{0.925} & \textbf{0.074} && \textbf{21.38} & \textbf{0.925} & \textbf{0.066} \\
\bottomrule
\end{tabular}
\caption{
\textbf{Ablations on Architecture and Flow Encoding.} The flow encoding we proposed in \cref{sec:arc_drag_conditions}, when injected into every LDM block at different resolutions, enhances the model's performance, regardless of the underlying LDM architecture.
}%
\label{tab:compare-architecture}
\vspace{-0.2in}
\end{table}

\begin{table}[t]
\small
\setlength\tabcolsep{5pt}
\renewcommand{\arraystretch}{1.0}
\centering
\begin{tabular}{@{}l ccc c ccc@{}}
\toprule
\multicolumn{1}{c}{\multirow{2}{*}[0ex]{\textbf{Training Data}}} & \multicolumn{3}{c}{\textbf{I.D. \dataset}} && \multicolumn{3}{c}{\textbf{O.O.D. \dataset}} \\
\cline{2-4}
\cline{6-8}
\multicolumn{1}{c}{} & PSNR$\uparrow$ & SSIM$\uparrow$ & LPIPS$\downarrow$ && PSNR$\uparrow$ & SSIM$\uparrow$ & LPIPS$\downarrow$ \\
\midrule
w/o Domain Random. & 21.38 & 0.925 & \textbf{0.066} && 18.03 & 0.897 & 0.104 \\
w/ Domain Random. & \textbf{21.82} & \textbf{0.928} & \textbf{0.066} && \textbf{19.74} & \textbf{0.920} & \textbf{0.083} \\
\bottomrule
\end{tabular}
\caption{
\textbf{Ablations on Domain Randomization.} Training jointly on the random texture renderings (\cref{sec:generalization}) bridges the gap between in-domain performance and out-of-distribution robustness.
\vspace{-1em}
}%
\label{tab:compare-data}
\end{table}

\begin{figure}[tb!]
\includegraphics[trim={0px 0px 0px 0px}, clip, width=\linewidth]{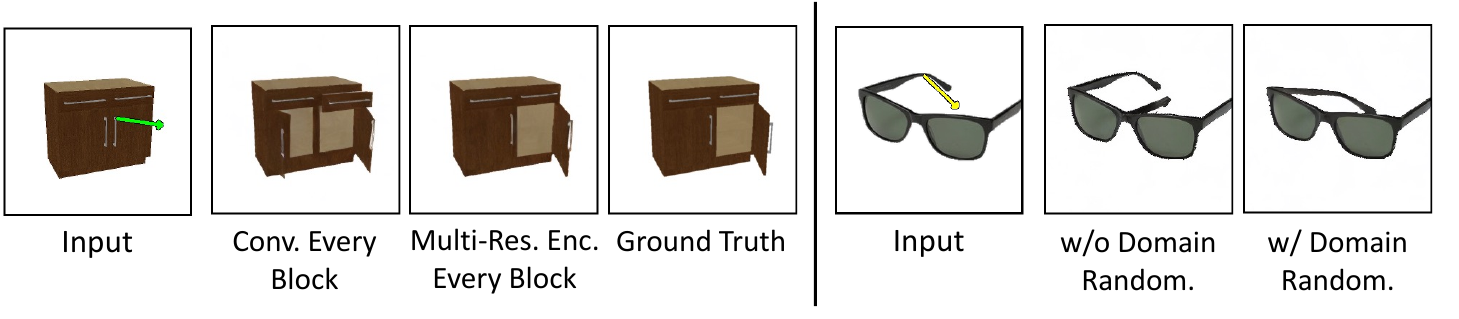}
\vspace{-2.0em}
\captionof{figure}{
\textbf{Qualitative Ablations.}
The visual results are consistent with the numerical ones in \cref{tab:compare-architecture,tab:compare-data} and validate our design choices.
\vspace{-1em}
}%
\label{fig:ablations}
\end{figure}

\subsection{Ablation Analysis and Limitations}

We present ablation studies on the two key components of \method: the multi-resolution flow encoding (\cref{sec:arc_drag_conditions}) and training with random texture (\cref{sec:generalization}).
As shown in \cref{tab:compare-architecture}, it is not sufficient to inject the drag conditioning through the input layer only (``Input Only'' in \cref{tab:compare-architecture}).
As previous works~\cite{yin2023dragnuwa, wang2023motionctrl} have discovered, injecting this motion conditioning at every LDM block (``Every Block'' in \cref{tab:compare-architecture}) is essential.
Our multi-resolution drag encoding better enforces disentanglement among different part motion than using a convolution network (\cref{fig:ablations} left).
It outperforms the previous methods and the improvement generalizes to both LDM architectures (\cref{tab:compare-architecture}).
Note also the importance of our domain randomization strategy (\cref{sec:generalization}).
Without training on the random texture renderings, the model's performance on out-of-distribution data drops significantly (\cref{tab:compare-data}), and generalizes poorly to unseen categories (\cref{fig:ablations} right).

\paragraph{Limitations.}

Currently, we do not explicitly enforce consistency of the generated images of the same object across different viewpoints and drag conditions.
We also train separate models for everyday objects and humans.
Extending a single model to all moving entities could help us obtain a universal motion prior.

\subsection{Application Results}

\begin{figure}[tb!]
\includegraphics[trim={2px 0px 0px 0px}, clip, width=\linewidth]{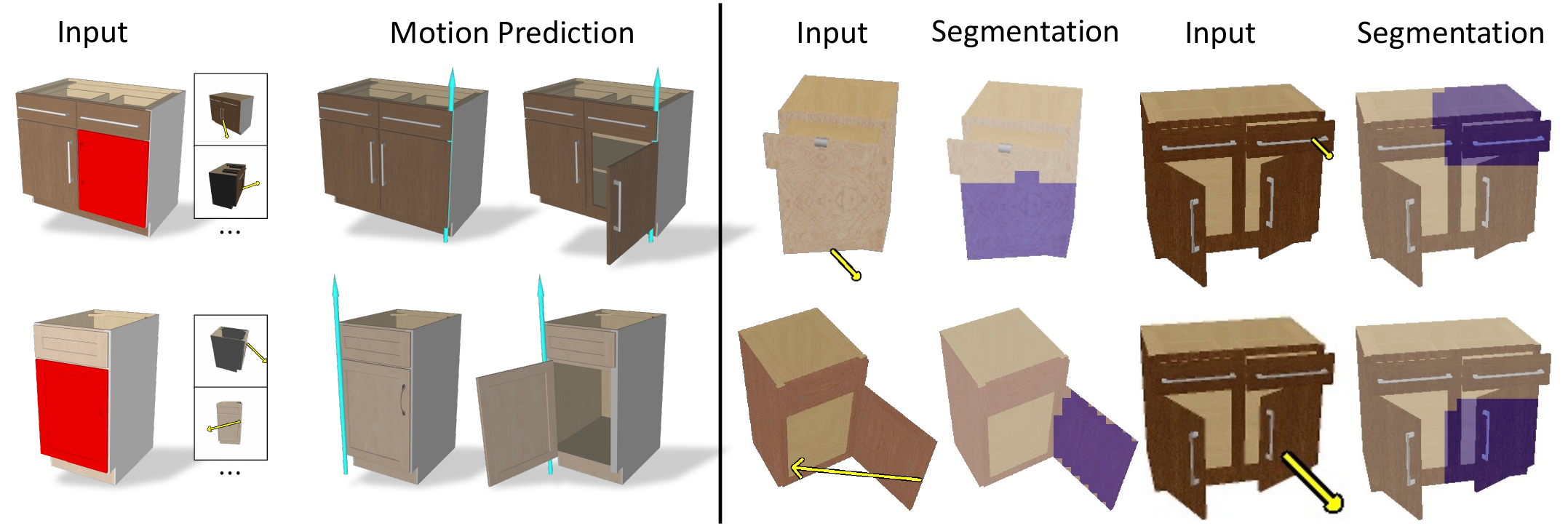}
\vspace{-1.5em}
\caption{
\method provides a useful motion model for \textbf{applications}, including motion analysis for articulated objects (left) and moving part segmentation (right).
\vspace{-1em}
}%
\label{fig:application-results}
\end{figure}

\cref{fig:application-results} (left) shows two motion analysis results, where we accurately recover the motion type and parameters specified by a drag.
This also indicates that our \method model is relatively consistent across different viewpoints.
In \cref{fig:application-results} (right) we show that \method's implicit part-level understanding can be leveraged to produce a coarse segmentation of moving parts.
Notably, the segmentation faithfully reflects the input drag.
In the sup.~mat., we provide a quantitative comparison of the segmentation quality on I.D.~\dataset (with available ground-truth masks) against baselines using CLIP~\cite{radford2021learning}, DINO~\cite{caron2021emerging, oquab2023dinov2} and SD~\cite{rombach2022stablediffusion}.

\section{Conclusion}%
\label{sec:conclusion}

We presented \method, an image generator that uses drags as an interface for \emph{part-level} dynamics.
By using a small amount of synthetic data, which we contributed, and domain randomization, \method can respond to drags by interpreting them as fine-grained part-level interactions with the underlying object.
Partly thanks to a new drag encoder, we have obtained better results than other methods on this task.
The model generalizes well to real-world images and novel categories.
We have also demonstrated some applications of the motion prior that \method provides.

% \paragraph{Ethics.}

% For further details on ethics, data protection, and copyright please see \url{https://www.robots.ox.ac.uk/~vedaldi/research/union/ethics.html}.

\paragraph{Acknowledgments.}
We thank Minghao Chen, Junyu Xie and Laurynas Karazija for insightful discussions and our sponsors Toshiba, ERC-CoG UNION 101001212 and EPSRC SYN3D EP/Z001811/1.

\bibliographystyle{splncs04}
\bibliography{main}

\clearpage
\appendix
\setcounter{table}{0}
\setcounter{figure}{0}
\renewcommand{\thetable}{\Alph{table}}
\renewcommand{\thefigure}{\Alph{figure}}

\section{Additional Details}%
\label{sec:supp-details}

\subsection{Implementation Details}%
\label{sec:supp-implement}

Here, we provide further architectural components in our implementation. This is an extension of \cref{sec:architecture}.

\paragraph{Injecting the Drag Encoding into DiT.}
Recall that our flow encoder $F_l$ produces an encoding of the drags $\mathcal{D}$ at the latent resolution of DiT, \ie, $F_l(\mathcal{D})\in \mathbb{R}^{C_l\times h\times w}$.
To inject this encoding to each DiT layer, the first step is to patchify $F_l(\mathcal{D})$ into a sequence of tokens, in line with how the noised latent is patchified.
Then, we regress another set of token-wise scale and shift parameters $\alpha, \gamma$, and $\beta$ to add on top of those regressed from the diffusion time step.

\paragraph{Injecting Global Image Control.}
Following the text-to-image diffusion model, we also inject additional image control into the network by using the CLIP encoding of the input image $y$, which provides global information.
Specifically, we leverage a pre-trained CLIP ViT/L-14 model to extract \texttt{[CLS]} tokens from every layer ($25$ in total) and train $25$ separate linear layers on top of the features.
The transformed features are then aggregated with average pooling and passed into another linear layer that produces the final image features.
These features are subsequently injected into the SD UNet by replacing the CLIP textual encoding, or into DiT by replacing the class label (as in our case there are neither text prompts nor class labels).

\subsection{Experiment Details}
\label{sec:supp-experiment}

\paragraph{Training Details.}
Our model was trained on $4\times$ NVIDIA A40 GPUs with batch size $4\times32$ and learning rate $10^{-5}$ for $150,000$ iterations using \texttt{AdamW} optimizer.
For the last $50,000$ iterations, we sample with probability $20\%$ random texture renderings and $80\%$ regular renderings in a batch.
For each batch item, we also zero out the drags $\mathcal{D}$ $10\%$ of the time to enable classifier-free guidance.

\paragraph{Inference Details.} All our samples are generated using $50$ denoising steps with classifier-free guidance weight $5$. Generating an image roughly takes 2 seconds on a single NVIDIA A40 GPU.

\paragraph{Evaluation Details.}
For quantitative evaluations, we compute all metrics (PSNR, SSIM and LPIPS) at resolution $256\times 256$.
However, \dragNUWA operates on a different aspect ratio (\ie, $576\times 320$).
For a fair comparison, we first pad the square input image $y$ along the horizontal axis to the correct aspect ratio and resize it to $576\times 320$, and then remove the padding of the generated last frame and resize it to $256\times 256$ for evaluation.
For methods that require a textual prompt (\dragdiffusion, \dragondiffusion), we used ``a rendering of a \texttt{[category]}'' wherein \texttt{[category]} is obtained from GAPartNet's categorization.
For \iptop, we generated the best possible results with the following prompts: ``pull the two drawers fully open'' (\cref{fig:qualitative-comp}a1), ``turn the bucket handle down to the left'' (\cref{fig:qualitative-comp}b1), ``pull the drawer directly beneath the table surface outward'' (\cref{fig:qualitative-comp}c1), and ``make every movable component articulate to its open position'' (\cref{fig:qualitative-comp}e1).

% \cref{fig:qualitative-comp}a1, b1, c1, e1):\begin{itemize}
%     \item ``Pull the two drawers fully open''.
%     \item ``Turn the bucket handle down to the left''.
%     \item ``Pull the drawer directly beneath the table surface outward''.
%     \item ``Make every movable component articulate to its open position''.
% \end{itemize}

\subsection{\dataset Dataset Details}
\label{sec:supp-dataset}

Here, we provide further details on the construction of our \dataset dataset.
This is an extension of \cref{sec:dataset}.

\paragraph{Motivation Behind the Selection Criteria for 3D Models.}
In GAPartNet~\cite{geng2023gapartnet}, some object components exhibit minimal or no movement, such as the keys on a remote or the handle on a teapot, which are \emph{not} useful for our objectives.
Heuristically, we find parts that are annotated with ``Hinge Handle'', ``Slider Drawer'', ``Hinge Lid'' or ``Hinge Door'' display substantial motion.
Hence, we discard objects that lack a minimum of one part annotated with these four kinematic types.
% GAPartNet is a collection of 3D models of everyday objects, annotated with part-level articulation information.
% Parts are labelled with their kinematic type (``Line Fixed Handle'', ``Round Fixed Handle'', ``Hinge Handle'', ``Hinge Lid'', ``Slider Lid'', ``Slider Button'', ``Slider Drawer'', ``Hinge Door'', ``Hinge Knob''), and the kinematic tree structure is annotated.
% The dataset contains 9 kinematic part types and 1,166 objects spanning 27 object categories with 8,489 parts annotated.
% Some objects, however, have parts with limited or no motion, such as the keys on a remote or the handle on  teapot, which are not useful for our purposes.
% Hence, we discard objects which do not have at least one part which is annotated with the one of the labels ``Hinge Handle'', ``Slider Drawer'', ``Hinge Lid'' and ``Hinge Door''.
% This yielded a set of $763$ candidate 3D models spanning $16$ categories.

\paragraph{Formulation of Asset Animation.} 
% The annotation of each object contains its kinematic tree structure, where nodes are rigid parts and edges are joints (fixed or not).
% Our goal is to sample a diversity of starting and ending states (before and after drag) of the different parts, including cases in which only a subset of parts move.

% Let $i$ denote the index of a part in the object.
For each part $i$, we define an articulation state
$
A \in \left[0, 1\right]
$,
where $A=0$ means fully closed and $A=1$ means fully open.
Because parts are essentially independent\footnote{With only a few exceptions where doors obstruct the movement of an internal part.}, we can sample any combination of states for the parts.
We construct the data by creating short ``animations'' consisting of $N+1=36$ frames.
For each animation, we start by first sampling a random subset $s$ of parts that move during the animation, whereas the rest are stationary.
Then, we generate two types of animations, selected with probability $p=0.5$.
For the first kind, the stationary parts $\bar s$ are fully closed, whereas the moving parts $s$ transition from fully closed to fully open, linearly.
For the second kind, each stationary part $i \in \bar s$ is set to a randomly sampled state $c_i$, whereas each moving part $i \in s$ transitions from a random starting state $a_i$ to a random ending state $b_i$.
Formally, the articulation state
$
A_{in}
$
of moving part $i$ at time $n \in \{0,\dots, N\}$ is given by:
\begin{equation}
A_{in} =
\begin{cases}
\frac{n}{N} \mathbbm{1}_{\left \{ i\in s \right \}} & \text{if the animation is of type 1} \\
c_i \mathbbm{1}_{\left \{ i\notin s \right \}} + \frac{(N-n)a_i + n b_i}{N}\mathbbm{1}_{\left \{ i\in s \right \}}  & \text{if the animation is of type 2} \\
\end{cases}
\end{equation}
where $a_i\sim U(0, \frac{1}{4}), b_i\sim U(\frac{3}{4}, 1)$ and $c_i\sim U(0, 1)$.

While in this formulation, each animation only contains parts opening, during training we randomly sample two frames, one as the reference image and the other as the ground truth image to be noised, irrespective of the order.

\subsection{Details on Drag-Driven Moving Part Segmentation}
\label{sec:supp-segmentation}
\paragraph{Feature Extraction.} 
Following existing works on semantic segmentation using image generative models, the first step is to extract the diffusion model's internal features.
Recall that at inference time, the model takes an image $y$ (to be segmented) and drags $\mathcal{D}$ as input and generates another image $x\sim \mathbb{P}(x|y,\mathcal{D})$ from pure Gaussian noise.
Formally, given an input image-drags pair $(y, \mathcal{D})$, we first sample a noisy latent $z_t$ at time step $t$ as:\begin{equation}
    z_t = \sqrt{1 - \sigma_t^2}z + \sigma_t \epsilon
\end{equation}
where $z = E(y)$ is the latent code of $y$ and $\epsilon\sim \mathcal{N}(0,\mathbf{I})$ is the added noise. We encode $\mathcal{D}$ using our proposed flow encoder $F_l$ and extract the diffusion internal features $f_l$ for the pair $(y, \mathcal{D})$ by feeding it into the UNet at block $l$:\begin{equation}
    f_l = \text{UNet}(z_t, F_l(\mathcal{D})) - \text{UNet}(z_t, F_l(\phi)).
\end{equation}
% Similarly, we compute the features without given the drags $\mathcal{D}$:\begin{equation}
%     f_l^{\phi} = \text{UNet}(z_t, F_l(\phi))
% \end{equation}
where $\phi$ represents no drags. Note we compute the feature \emph{difference} here to better reflect the effect of the drags.

In our implementation, the drag conditions are injected into the self-attention layer of each UNet block.
Empirically, we find using the output of the self-attention layer instead of the whole UNet block, provides better segmentation results.
Following previous works~\cite{baranchuk2021label, xu2023open}, we resize the features at each block $l$ to the image resolution and concatenate them to create a feature pyramid.

\paragraph{Clustering.}
We perform K-means clustering on the feature pyramid to group the object's foreground pixels.
We then obtain the segmentation by using the clusters to which the drag tails are associated.
\newline\newline
We treat the time step $t$ used for feature extraction and the number of clusters $N_c$ as hyperparameters. An analysis of their effect is given in \cref{sec:compare-seg}.

\section{Additional Results}
\label{sec:supp-results}

\subsection{Additional Qualitative Results}
\label{sec:addition-results}
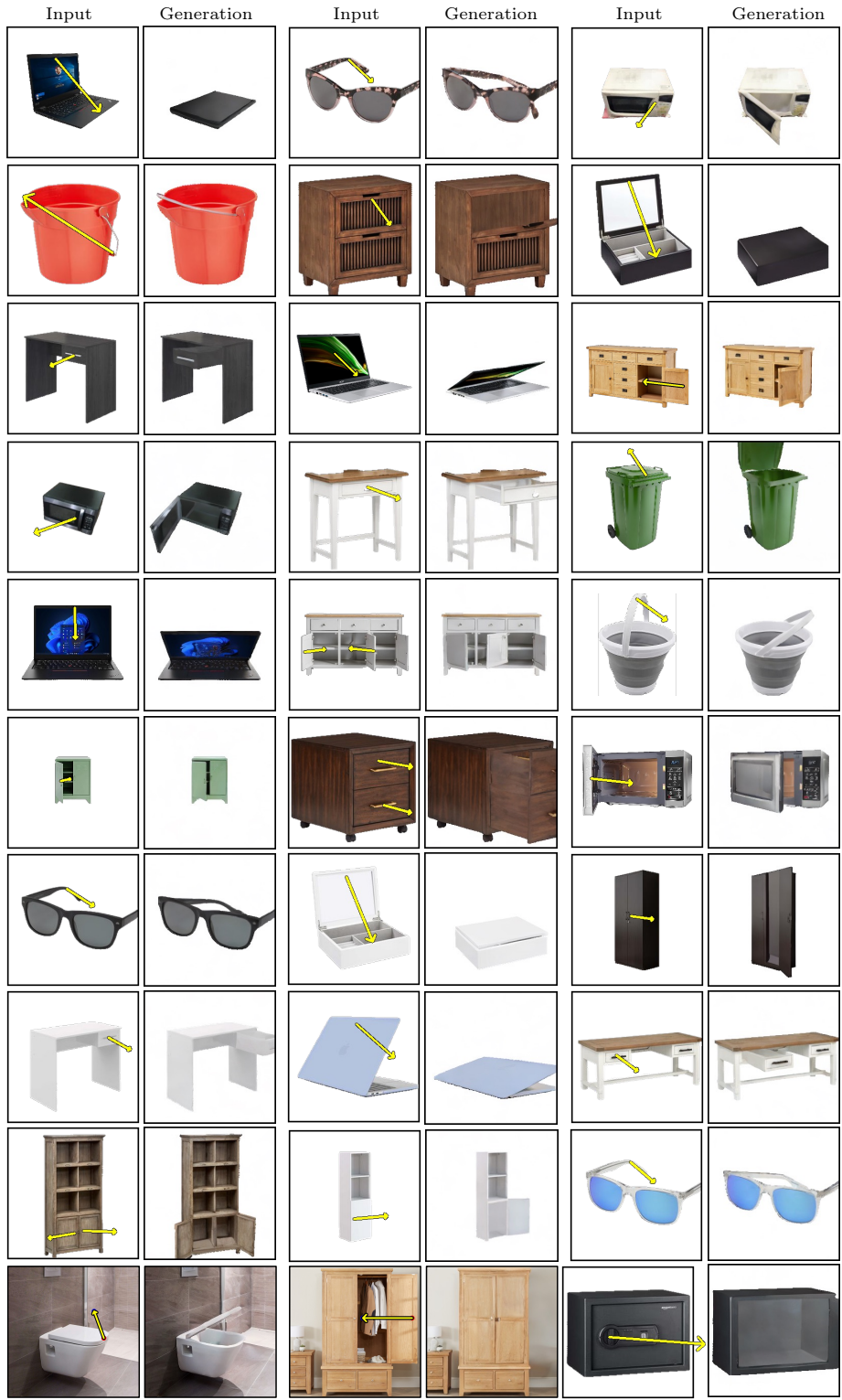
\begin{figure}[tb!]
\includegraphics[trim={0px 0px 0px -2px}, clip, width=\linewidth]{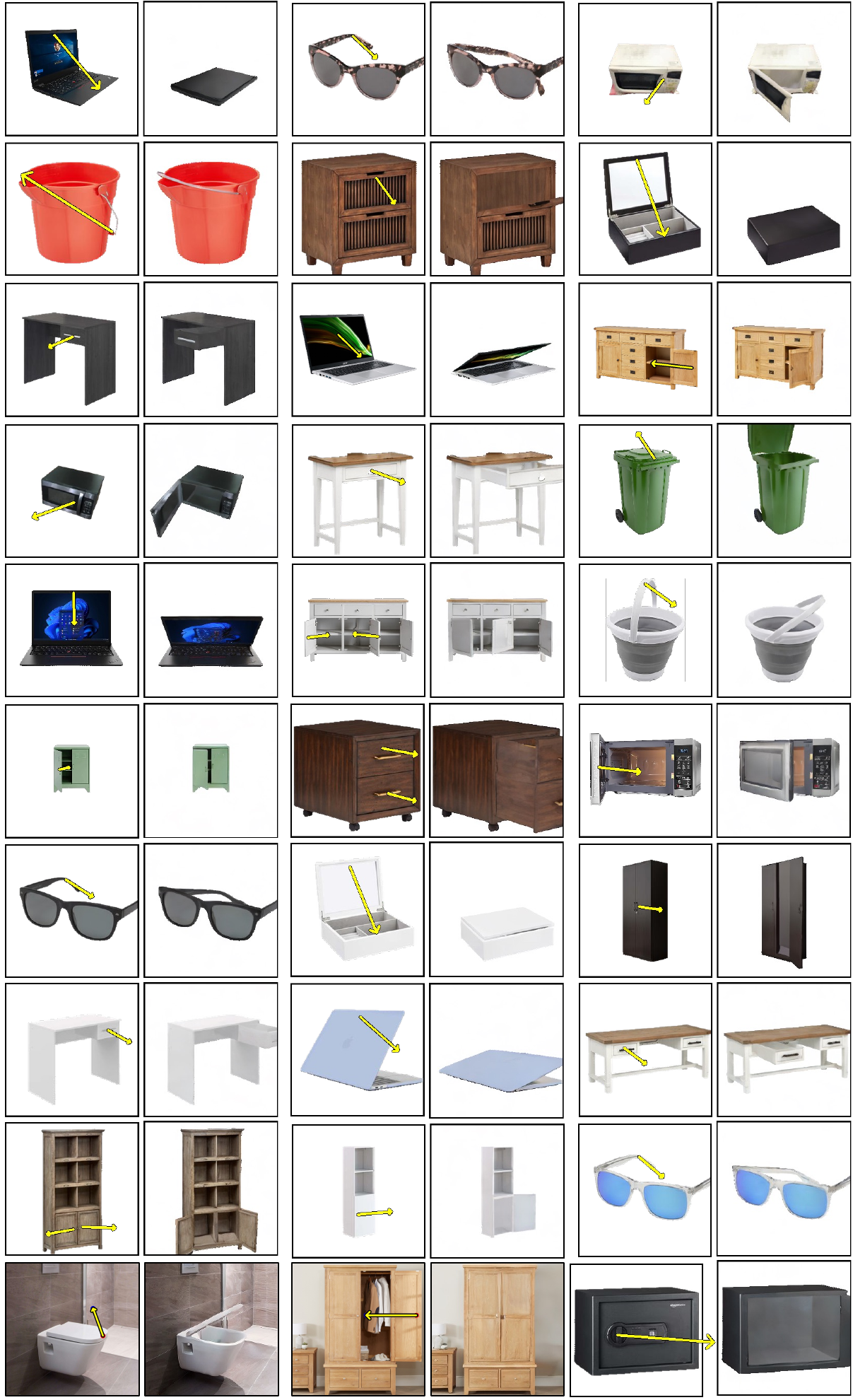}
\begin{picture}(0,0)
    \put(18,583){\scriptsize Input}
    \put(65,583){\scriptsize Generation}
    \put(137,583){\scriptsize Input}
    \put(183,583){\scriptsize Generation}
    \put(254,583){\scriptsize Input}
    \put(302,583){\scriptsize Generation}
\end{picture}
\vspace{-1.5em}
\caption{
\textbf{Additional Qualitative Results.}
}%
\label{fig:additional-results}
\end{figure}

Additional results on a variety of categories can be found in \cref{fig:additional-results}.
In instances where the conditional image features a complex, real-world background,
the model adeptly ``fills in'' the occluded background details. This is notable given that its training exclusively utilizes images with white backgrounds. This capability is largely thanks to the strong priors of the pre-trained SD model.
% Note that the model generates realistic images with diverse, real-world backgrounds, even though it is solely trained with images of white background, thanks to the prior captured by the pre-trained SD.

\subsection{Failure Cases}
\label{sec:failure-cases}
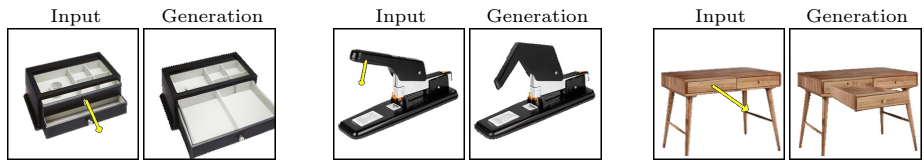
\begin{figure}[tb!]
\includegraphics[trim={0px 0px 0px 0px}, clip, width=\linewidth]{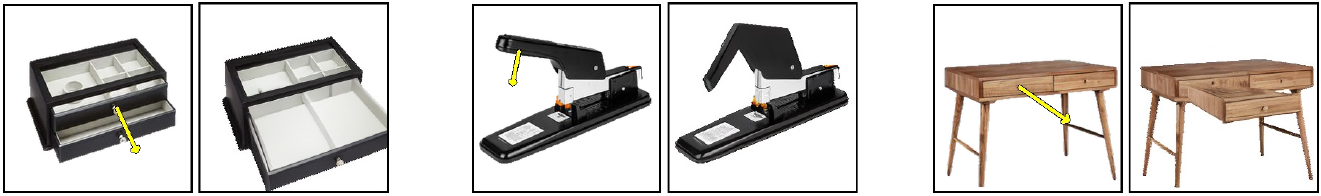}
\begin{picture}(0,0)
    \put(17,64){\scriptsize Input}
    \put(59,64){\scriptsize Generation}
    \put(140,64){\scriptsize Input}
    \put(182,64){\scriptsize Generation}
    \put(261,64){\scriptsize Input}
    \put(303,64){\scriptsize Generation}
\end{picture}
\vspace{-1.5em}
\caption{
\textbf{Failure Cases.} While our model is capable of generating images of complex part-level dynamics, it occasionally produces physically implausible samples.
}%
\label{fig:failure-cases}
\end{figure}
\begin{table}[t]
\small
\setlength{\tabcolsep}{0.1cm}
\renewcommand{\arraystretch}{1.0}
\centering
\begin{tabular}{@{}cccccc c ccc@{}}
\toprule
\multicolumn{1}{c}{\multirow{2}{*}[0ex]{\makecell{\textbf{\# of Clusters} \\ $N_c$}}} &
\multicolumn{5}{c}{\textbf{\method Features}} & &
\multicolumn{3}{c}{\textbf{Baseline Features}} \\
\cline{2-6}
\cline{8-10}
& $t=0$ & $t=200$ & $t=500$ & $t=800$ & $t=999$ & & CLIP & DINO & SD \\
\midrule
2 & 22.79 & 22.71 & 21.72 & 23.32 & \textbf{23.39} && 13.89 & 21.24 & 14.47 \\
3 & 26.84 & \textbf{27.22} & 26.94 & 26.42 & 25.81 && 14.00 & 24.13 & 14.38 \\
4 & 26.06 & \underline{\textbf{27.29}} & 26.45 & 26.62 & 26.40 && 14.49 & 24.16 & 14.68 \\
5 & 25.73 & 25.48 & 25.21 & 26.02 & \textbf{26.10} && 13.55 & 24.58 & 14.62 \\
\bottomrule
\end{tabular}
\caption{
\textbf{Quantitative Evaluation of Drag-Driven Moving Part Segmentation.} Clustering of our \method features yields the best (\ie, highest) mIoU, and is relatively robust across different hyperparameter settings.
}%
\label{tab:compare-seg}
\vspace{-0.2in}
\end{table}
We show typical failure cases of our model in \cref{fig:failure-cases}.
Notably, the model's limitations, potentially attributed to the limited scale of the training data, manifest in several ways:
it may i) incorrectly recognize different motion parts of an object (left);
ii) unrealistically distort an object from an \emph{unseen} category (middle);
or iii) occasionally create images that lack physical plausibility (right).

\subsection{Quantitative Comparisons of Moving Part Segmentation}
\label{sec:compare-seg}
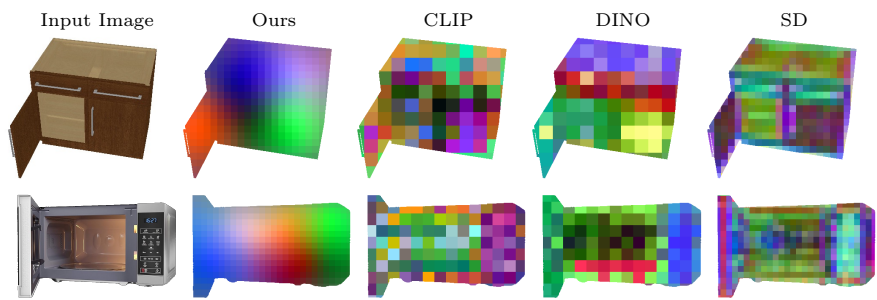
\begin{figure}[tb!]
\includegraphics[trim={0px 0px 0px -25px}, clip, width=\linewidth]{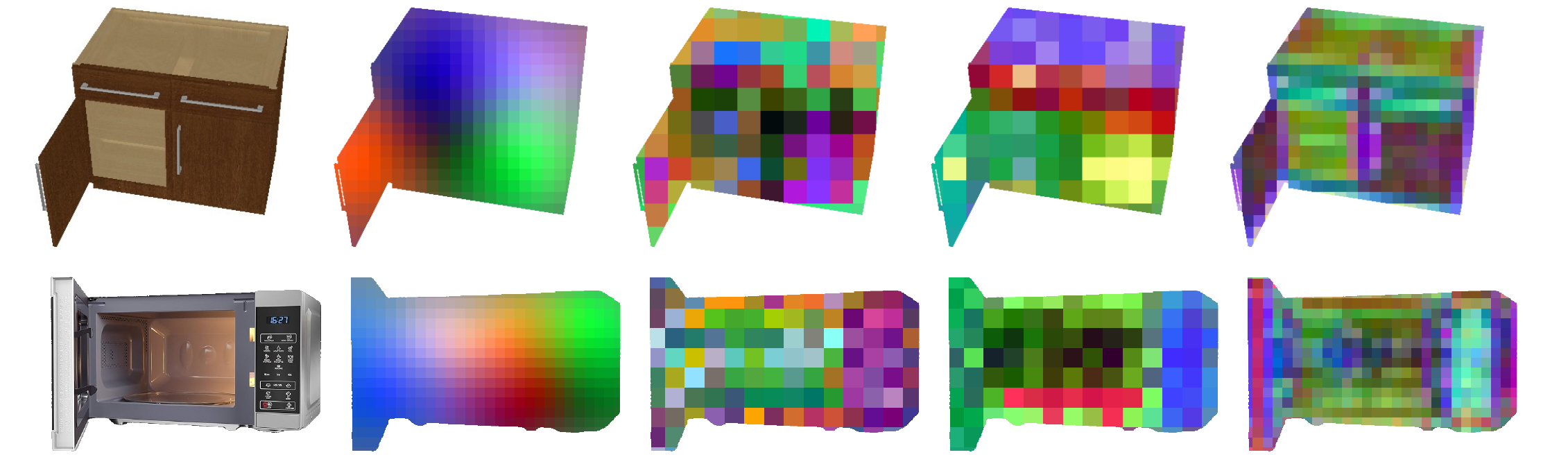}
\begin{picture}(0,0)
    \put(20,115){\scriptsize Input Image}
    \put(100,115){\scriptsize Ours}
    \put(165,115){\scriptsize CLIP}
    \put(230,115){\scriptsize DINO}
    \put(300,115){\scriptsize SD}
    % \put(0,83){\footnotesize a)}
    % \put(118,83){\footnotesize b)}
    % \put(238,83){\footnotesize c)}
    % \put(0,33){\footnotesize d)}
    % \put(118,33){\footnotesize e)}
    % \put(238,33){\footnotesize f)}
\end{picture}
\vspace{-1.5em}
\caption{
\textbf{Visualization of Image Features.} Notably, the features extracted through our \method are smooth and exhibit part-level information.
}%
\label{fig:feature-pca}
\end{figure}
We assess our proposed moving part segmentation approach on the test split of \dataset,
where part-level segmentation is annotated so the ground-truth masks are available.
In \cref{tab:compare-seg}, we provide the average mask Intersection over Union (mIoU) obtained using our method and baselines of clustering alternative CLIP~\cite{radford2021learning}, DINO~\cite{caron2021emerging, oquab2023dinov2} and SD\footnote{To ensure fairness, the metric for SD is obtained by selecting the maximum from the $5$ mIoU values with diffusion time step set to $0$, $200$, $500$, $800$ and $999$ respectively.}~\cite{rombach2022stablediffusion} features.
We also visualize the corresponding features (transformed through principal component analysis on foreground pixels to $3$ channels as RGB values) in \cref{fig:feature-pca}.
The results underscore that the features extracted through our \method demonstrate better part-level information.

\end{document}